\documentclass[11pt]{article}

\usepackage[margin=2.5cm]{geometry}

\usepackage{xcolor}

\usepackage{amsmath}
\usepackage{amssymb}
\usepackage{newunicodechar}
\newunicodechar{↻}{\ensuremath{\circlearrowright}}

\usepackage{graphicx}
\usepackage[percent]{overpic}

\usepackage{booktabs}
\usepackage{multirow}
\usepackage{longtable}
\usepackage{tabularx}
\usepackage{array}

\usepackage{calc}
\usepackage{pifont}
\usepackage{etoolbox}
\usepackage{authblk}
\usepackage{xurl}

\usepackage{fontspec}

\IfFileExists{microtype.sty}{
	\usepackage{microtype}
	\UseMicrotypeSet[protrusion]{basicmath}
}{}

\IfFileExists{upquote.sty}{
	\usepackage{upquote}
}{}

\makeatletter
\patchcmd{\longtable}
{\par}
{\if@noskipsec\mbox{}\fi\par}
{}
{}
\makeatother

\IfFileExists{footnotehyper.sty}{
	\usepackage{footnotehyper}
}{
	\usepackage{footnote}
}

\makesavenoteenv{longtable}

\makeatletter

\newsavebox{\pandoc@box}

\newcommand*\pandocbounded[1]{%
	\sbox{\pandoc@box}{#1}%
	\Gscale@div\@tempa
	{\textheight}
	{\dimexpr\ht\pandoc@box+\dp\pandoc@box\relax}%
	\Gscale@div\@tempb
	{\linewidth}
	{\wd\pandoc@box}%
	\ifdim\@tempb\p@<\@tempa\p@
	\let\@tempa\@tempb
	\fi
	\ifdim\@tempa\p@<\p@
	\scalebox{\@tempa}{\usebox{\pandoc@box}}%
	\else
	\usebox{\pandoc@box}%
	\fi
}

\def\fps@figure{htbp}

\makeatother

\IfFileExists{xurl.sty}{
	\usepackage{xurl}
}{}

\usepackage[
backend=biber,
style=ieee,
sorting=none
]{biblatex}

\DeclareFieldFormat{doi}{%
	\mkbibacro{DOI}\addcolon\space
	\ifhyperref
	{\href{https://doi.org/#1}{\mbox{\nolinkurl{#1}}}}
	{\mbox{\nolinkurl{#1}}}%
}

\DeclareSourcemap{
	\maps[datatype=bibtex]{
		\map{
			\pertype{inproceedings}
			\step[fieldsource=eventtitle, fieldtarget=booktitle]
			\step[fieldset=eventtitle, null]
		}
	}
}

\AtEveryBibitem{%
	\ifentrytype{inproceedings}
	{%
		\clearfield{url}%
		\clearfield{urlyear}%
		\clearfield{urlmonth}%
		\clearfield{urlday}%
	}
	{%
		\iffieldundef{doi}
		{}
		{%
			\clearfield{url}%
			\clearfield{urlyear}%
			\clearfield{urlmonth}%
			\clearfield{urlday}%
		}%
	}%
}

\usepackage[unicode,hidelinks]{hyperref}
\usepackage{bookmark}

\hypersetup{
	pdftitle={
		DreamFly: Causal Memory and Receding-Horizon Diffusion Planning
		for Aerial Vision-Language Navigation
	},
	pdfauthor={Yan Deng and Fei Xu}
}

\title{
	DreamFly: Causal Memory and Receding-Horizon Diffusion Planning
	for Aerial Vision-Language Navigation
}

\author[1]{Yan Deng\thanks{Corresponding author: dengyan@st.xatu.edu.cn}}
\author[2]{Fei Xu}

\affil[1]{
	School of Electronic Information Engineering,
	Xi’an Technological  University,
	Xi’an, China
}

\affil[2]{
	School of Computer Science and Engineering,
	Xi’an Technological  University,
	Xi’an, China
}

\date{}

\begin{document}
\maketitle

\begin{abstract}
	
	Aerial vision-language navigation (VLN) requires an embodied agent to integrate visual evidence over time, plan future actions, and determine when it has reached a navigation goal under partial observability. Although recent VLA models offer a promising perception-to-action paradigm, adapting them to aerial navigation remains challenging. Policies conditioned primarily on the current observation may lose track of previously observed landmarks; single-step action prediction provides limited lookahead; and implicit termination through action generation may be unreliable.
	
	To address these challenges, we propose DreamFly, a diffusion-based aerial VLN framework built on Dream-VLA. DreamFly first introduces a causally aligned historical memory that augments the current visual representation with evidence drawn exclusively from observations preceding the current decision step, thereby enabling temporal reasoning without leaking future information. Leveraging the bidirectional diffusion backbone, we further formulate navigation as receding-horizon diffusion planning, where the policy jointly predicts a $K$-step action chunk but executes only the first action before replanning from the next observation. This plan-$K$, execute-one strategy treats predicted future actions as auxiliary planning targets while preserving closed-loop visual feedback. Finally, we introduce LiteStop, a lightweight termination module that estimates the stop probability directly from the action logits at the initial all-mask state, thereby decoupling explicit termination from action generation.
	
	Together, these components establish a causal closed-loop cycle of observation, memory retrieval, diffusion planning, termination assessment, execution, and memory update. Experiments on the OpenFly benchmark demonstrate consistent improvements in both seen and unseen environments. DreamFly achieves 32.04\%/29.46\% SR and 28.22\%/23.54\% SPL on the test-seen/test-unseen splits, respectively, outperforming all compared methods on both metrics while also attaining the lowest navigation error. These results highlight the effectiveness of jointly modeling historical context, future action structure, and explicit termination for aerial vision-language navigation.
\end{abstract}


\section{Introduction}
\label{sec:introduction}

Vision-and-Language Navigation (VLN) requires embodied agents to interpret natural-language instructions and navigate to specified goals using sequentially acquired visual observations \cite{anderson2018}. In contrast to static visual recognition and single-step action classification, VLN is inherently a partially observable, closed-loop sequential decision problem: an action taken at the current step affects the agent's subsequent state and visual observation, while newly acquired environmental evidence updates its estimates of task progress, spatial context, and direction of travel. A reliable VLN policy therefore requires more than vision--language grounding at an isolated decision step. It must maintain temporal consistency across historical visual evidence, the current observation, planned future actions, and executed actions throughout its interaction with the environment.

Aerial vision-language navigation extends VLN to unmanned aerial vehicles (UAVs), requiring agents to navigate autonomously through large-scale 3D environments by following natural-language instructions \cite{AerialVLN}. However, aerial VLN cannot be treated as a direct transfer of conventional ground-based VLN to a flying platform. Ground agents typically move on approximately two-dimensional traversable surfaces, whereas UAVs must jointly coordinate horizontal displacement, vertical motion, and viewpoint adjustment. Changes in altitude affect not only the UAV's spatial position but also the spatial extent of the visible scene, the apparent scale of landmarks, the level of visible detail, and obstacle visibility, thereby tightly coupling horizontal and vertical navigation decisions \cite{Grid-basedViewSelection}. Moreover, urban aerial environments often span large areas and contain complex 3D structures and substantial visual occlusion, such that a single egocentric observation captures only a limited portion of the surrounding scene. Consequently, local decision errors can compound over time. These characteristics impose stringent requirements on historical context modeling, 3D spatial reasoning, cross-modal grounding, and online error correction in aerial VLN \cite{AerialVLN,Grid-basedViewSelection,OpenFly}.

Early VLN methods commonly relied on sequence-to-sequence architectures, typically implemented as recurrent policies, to autoregressively predict the next navigation action from the language instruction, current visual observation, and recurrent hidden state \cite{anderson2018,wangReinforcedCrossModalMatching2019}. With advances in vision--language pretraining and Transformer-based architectures, subsequent studies developed recurrent vision--language Transformers, history-aware representations, and explicit environmental memories to support reasoning across successive decision steps. For example, Recurrent VLN-BERT maintains a recurrent state representation, whereas HAMT employs a hierarchical Transformer to encode navigation history, thereby facilitating multimodal reasoning over extended trajectories \cite{RecurrentVLNBERT,chenHistoryAwareMultimodal2021}. In aerial VLN, recent approaches have further explored multidirectional view selection, bird's-eye-view feature maps, keyframe selection, and compression of historical visual observations to mitigate restricted egocentric visibility, redundant observations, and difficulties in tracking navigation state \cite{Grid-basedViewSelection,OpenFly,LongFly}. Together, these studies underscore the importance of historical information for navigation. However, historical modeling depends not only on \emph{how} past observations are represented, but also on precisely \emph{when} they become available to the policy during closed-loop interaction.

More recently, vision--language--action (VLA) models have emerged as a promising paradigm for language-conditioned end-to-end control. By formulating actions as discrete tokens or continuous values, VLA models adapt pretrained vision--language models to generate robot actions conditioned on language instructions and visual observations \cite{RT2,OpenVLA}. Action chunking captures short-horizon temporal structure by jointly predicting a sequence of actions, while diffusion-based policies model multimodal action distributions through iterative denoising and can be deployed with receding-horizon replanning \cite{chiDiffusionPolicyVisuomotor2025}. Dream-VLA, for example, employs a diffusion Transformer to jointly predict action chunks, thereby capturing dependencies among actions over a short future horizon \cite{zhangDreamVLAVisionLanguageActionModel2025}.
Recent studies of aerial navigation have likewise explored end-to-end VLA policies, diffusion-based action modeling, and joint world--action prediction, highlighting the potential of explicitly predicting future states or action sequences for navigation in complex 3D environments \cite{xuAerialVLAVisionLanguageActionModel2026b,WorldVLN,ImagineUAV,FSDVLN}. Several of these approaches seek to enhance planning by forecasting future world states or visual observations. However, how to exploit dependencies among predicted future actions over a short horizon while retaining closed-loop replanning from newly acquired observations remains underexplored.

Despite recent advances in historical modeling, 3D reasoning, and multi-step action prediction, three temporal aspects of aerial VLN remain insufficiently addressed within a unified closed-loop decision process. First, an online historical memory requires an explicit temporal boundary. Here, \emph{causal} refers to temporally ordered information access rather than causal relations between variables: at decision step $t$, historical information is restricted to observations acquired before step $t$. Accordingly, the historical memory $M_{<t}$ is read before the current observation is written to it, preventing current-step information from entering the historical branch. Applying this read-before-write ordering during both training and deployment gives $M_{<t}$ a well-defined temporal boundary and ensures consistent information availability across the two phases. This temporal notion of causality should not be conflated with covariate shift between expert and policy-induced state distributions in imitation learning \cite{rossReductionImitationLearning2011}. It does not mitigate the covariate shift arising from policy execution; rather, it constrains when observations enter the historical memory.

Second, multi-step action prediction should provide lookahead for the current decision without sacrificing closed-loop execution. Replanning after each executed action allows the agent to exploit dependencies among predicted future actions without committing to an open-loop action chunk, thereby retaining responsiveness to newly acquired observations \cite{chiDiffusionPolicyVisuomotor2025}. Third, termination carries markedly different consequences from ordinary motion actions. Motion errors can often be corrected through subsequent actions, whereas a premature stop irreversibly ends the navigation episode. Treating termination and motion actions under the same prediction and calibration mechanism can therefore obscure the asymmetric risks of premature and delayed stopping \cite{xiangLearningStopSimple2020}. Together, these considerations motivate the explicit coordination of historical information, future-action planning, and termination within a temporally consistent closed-loop decision process.

To address these issues, we propose DreamFly, a temporally consistent, closed-loop aerial VLN framework built on discrete diffusion-based action generation. First, DreamFly introduces a causally aligned historical memory constructed exclusively from visual observations acquired before the current decision step. Under a read-before-write protocol, $M_{<t}$ is held fixed during decision step $t$ and updated with the current observation only after the current decision, for use at subsequent steps. The resulting historical representation is fused with the current visual features through a lightweight memory-conditioning module, allowing the policy to reason jointly over the language instruction, historical visual evidence, and current observation.

Building on the memory-conditioned representation, DreamFly performs receding-horizon diffusion planning by jointly predicting a chunk of $K$ discrete actions that comprises the current action and captures dependencies among actions over a short future horizon. During online inference, DreamFly follows a \emph{plan-$K$, execute-one} strategy: it executes only the first action and replans upon receiving the next observation. In this way, the future positions in the action chunk serve as auxiliary prediction targets for the current decision, while execution remains closed loop.

\begin{figure}[t]
	\centering
	\includegraphics[width=\linewidth]{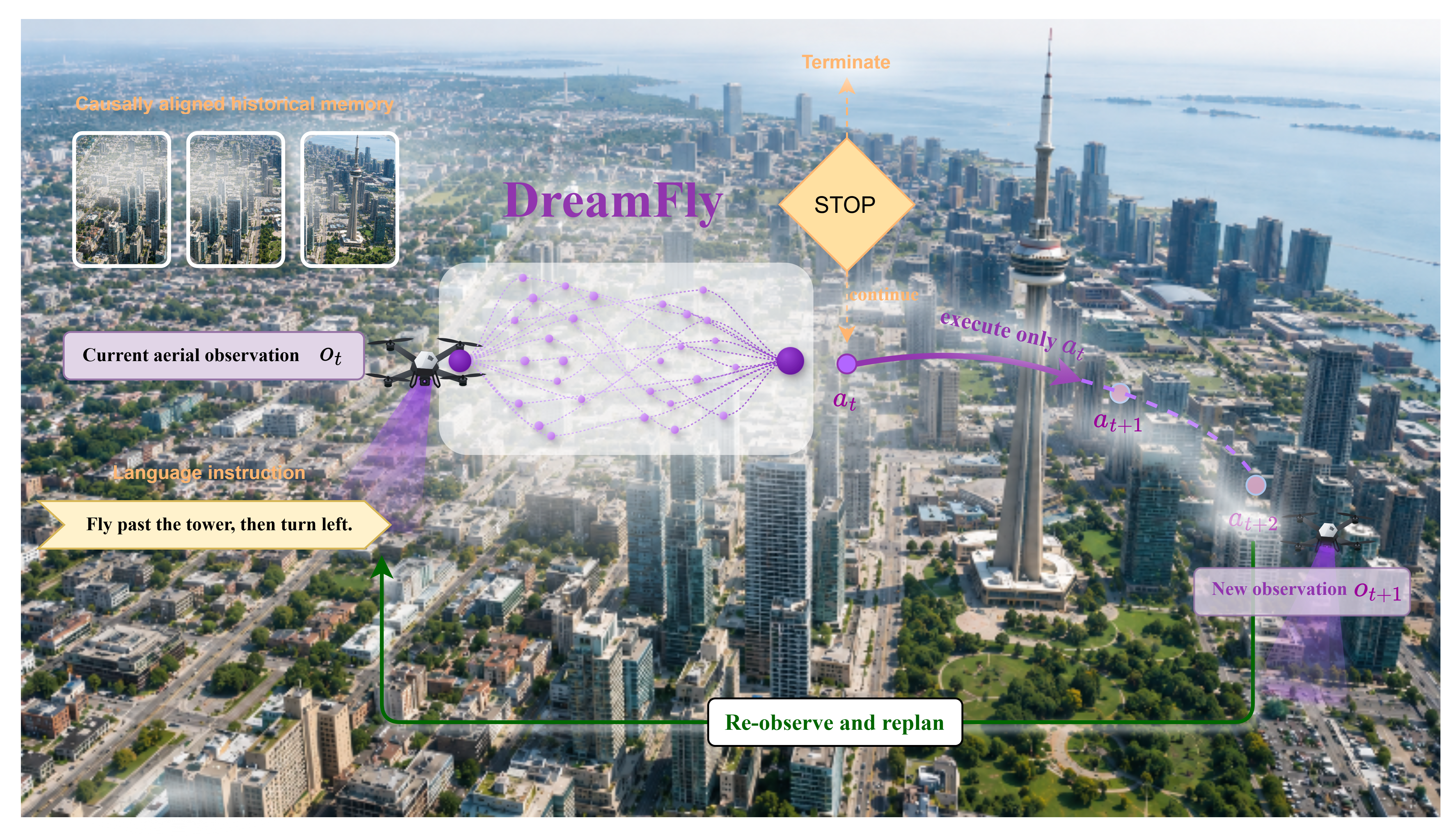}
	\caption{DreamFly at a Glance. At decision step $t$, DreamFly conditions on the language instruction, current aerial observation, and a historical memory containing only observations acquired before step $t$ to generate a $K$-step action plan. DreamFly executes only the first action, acquires a new observation, and replans at the next decision step, while LiteStop separately estimates the probability of termination. Dashed actions denote planned future actions that are not executed at the current decision step.}
	\label{fig:teaser}
\end{figure}

Finally, we introduce LiteStop, a lightweight module that explicitly predicts termination. Rather than treating termination as an ordinary motion action, LiteStop estimates the probability of stopping from the action-token logit grid $\mathbf H_t^{(0)}$ extracted from the diffusion policy at its initial all-mask state. LiteStop is trained separately while the navigation policy remains frozen, providing a dedicated termination objective without modifying the learned motion policy. Together, causally aligned historical memory, receding-horizon diffusion planning, and explicit termination form a temporally consistent, closed-loop navigation process, as illustrated in Fig.~\ref{fig:teaser}.

We make the following three main contributions:

\begin{enumerate}
	\item \textbf{We introduce a causally aligned historical memory for aerial VLN with an explicit temporal information boundary.}
	DreamFly constructs its historical memory exclusively from observations acquired before the current decision step and adopts a read-before-write protocol that prevents current-step information from entering the historical branch. A lightweight gated cross-attention module selectively integrates task-relevant historical evidence into the current visual representation, thereby supporting temporally consistent reasoning under partial observability.
	
	\item \textbf{We propose receding-horizon diffusion planning that exploits dependencies among predicted future actions while preserving closed-loop control.}
	DreamFly jointly predicts the current action and a short sequence of future actions as a chunk of $K$ discrete actions, using valid-prefix supervision and horizon-aware loss weighting to emphasize the immediately executable action. During inference, DreamFly executes only the first action and replans upon receiving the next observation, allowing the predicted future actions to serve as auxiliary planning targets while preserving closed-loop execution.
	
	\item \textbf{We develop LiteStop, an explicit and decoupled termination module for aerial VLN.}
	Rather than coupling termination with motion-action generation, LiteStop estimates the probability of stopping from the action logits produced by the diffusion policy at its initial all-mask state and is trained separately while the navigation policy remains frozen. This decoupling provides a dedicated termination objective without modifying the learned motion policy.
	
\end{enumerate}

\section{Related Work}
\label{sec:related-work}

\subsection{Aerial Vision-Language Navigation}
\label{sec:related-aerial-vln}

Aerial VLN requires a UAV to navigate toward specified goals in large-scale 3D environments by following natural-language instructions and relying on sequentially acquired egocentric observations. Early studies focused primarily on task formulation and benchmark construction. AVDN introduced an interactive UAV navigation setting based on asynchronous multi-round dialogue, enabling the agent to request and use additional linguistic guidance during navigation \cite{fanAerialVisionandDialogNavigation2023}. AerialVLN subsequently established a setting for instruction-guided UAV navigation in continuous urban environments, together with systematic data-collection and evaluation protocols \cite{AerialVLN}. Subsequent benchmarks have progressively extended aerial VLN to larger-scale and more realistic settings. CityNav extends aerial navigation to city-scale target localization using real-world urban aerial data and a combination of visual and geographic cues \cite{leeCityNavLargeScaleDataset2025}. OpenUAV incorporates continuous flight trajectories, realistic flight dynamics, and complex task scenarios, providing a setting closer to practical UAV control than discrete simulated navigation \cite{wangRealisticUAVVisionLanguage2024}. OpenFly introduces an automated data-generation pipeline that integrates rendered and real-world scenes, thereby broadening the range of training trajectories and visual domains \cite{OpenFly}. More recently, AirNav constructs a large-scale benchmark from real-world urban aerial data paired with diverse natural-language navigation instructions \cite{caiAirNavLargeScaleUAV2026}. Collectively, these efforts have expanded aerial VLN from early simulated formulations to continuous, large-scale, and increasingly realistic navigation settings.

Building on these benchmarks, recent aerial VLN methods have primarily pursued two complementary directions: structured spatial and historical modeling, and reasoning with foundation models. One line of work explicitly structures spatial relations and observation histories to support navigation in large-scale environments. STMR projects instruction-relevant visual semantics into a semantic--topological--metric representation and uses a large language model to reason over this representation and predict navigation actions \cite{gaoExploringSpatialRepresentation2025}. OpenFly-Agent selects informative keyframes from long observation histories to provide compact visual context for each navigation decision \cite{OpenFly}. CityNavAgent uses hierarchical semantic planning to decompose complex instructions into subgoals at multiple levels of granularity while maintaining a global topological memory of explored trajectories \cite{leeCityNavLargeScaleDataset2025}. HETT combines a historical grid representation with a two-stage, coarse-to-fine decision process that links global target localization to local action selection \cite{dingHistoryEnhancedTwoStageTransformer2025}. LookasideVLN incorporates directional semantics and lookahead path reasoning to improve landmark selection and spatial reasoning in complex urban environments \cite{LookasideVLN}. Collectively, these methods demonstrate the value of structured spatial representations and accessible navigation histories for multi-step decision making in large-scale aerial environments.

A complementary line of work leverages vision-language foundation models and large language models to enhance cross-modal understanding and high-level reasoning. The navigation model accompanying OpenUAV employs a multimodal language model to jointly process multi-view observations, task descriptions, and auxiliary information, generating flight trajectories hierarchically \cite{wangRealisticUAVVisionLanguage2024}. FlightGPT combines supervised fine-tuning with reinforcement learning to strengthen vision-language reasoning and navigation decision making \cite{caiFlightGPTGeneralizableInterpretable2025}. FineCog-Nav decomposes the navigation process into fine-grained modules for language understanding, perception, memory, imagination, and decision making, investigating how coordinated foundation-model reasoning can support zero-shot UAV navigation \cite{shaoFineCogNavIntegratingFinegrained2026a}. Overall, prior work has made substantial progress in data scale, spatial representation, historical-context modeling, and foundation-model reasoning. However, how to coordinate the distinct temporal roles of historical context, multi-step action planning, and navigation termination within a unified closed-loop decision process remains underexplored.

\subsection{Historical Context Modeling and Memory in VLN}
\label{sec:related-history-memory}

Because each visual observation captures only a partial view of the environment, VLN policies must use historical information to retain spatial context from earlier observations, account for prior actions, and track instruction-following progress. Early approaches commonly summarized navigation history in recurrent states. Recurrent VLN-BERT summarizes navigation history in a compact recurrent state within a cross-modal Transformer rather than retaining explicitly addressable representations of individual past observations \cite{RecurrentVLNBERT}. In contrast, HAMT uses a hierarchical Transformer to encode the full history of panoramic observations, retaining detailed temporal context, although the amount of historical input and potential redundancy can increase as the trajectory grows \cite{chenHistoryAwareMultimodal2021}. These approaches illustrate two different points in the trade-off between compact representation and detailed history preservation.

To support explicit storage and retrieval of past visual evidence, other studies developed spatially structured memory representations. Structured Scene Memory organizes visual and geometric cues acquired during navigation into an explicit scene representation and employs an adaptive collection-and-read mechanism to support reasoning over time \cite{wangStructuredSceneMemory2021}. GridMM projects historical observations onto a dynamically expanding egocentric bird's-eye-view grid and aggregates instruction-relevant visual evidence across spatial regions \cite{GridMM}. These approaches organize historical evidence through scene connectivity or relative spatial location, making spatial relations explicit for multi-step reasoning and planning. Their memory construction is consequently coupled with the spatial organization or projection mechanism adopted by the navigation policy.

The characteristics of aerial VLN, including long flight trajectories, frequent viewpoint changes, and substantial visual redundancy, have further motivated the development of compact representations of observation history. OpenFly-Agent employs adaptive frame-level token sampling to retain informative historical visual content while reducing redundancy among consecutive aerial observations \cite{OpenFly}. LongFly compresses historical images into a fixed-length context representation through slot-based aggregation and integrates the resulting representation with spatiotemporal trajectory encoding and prompt-guided multimodal fusion for navigation decision-making \cite{LongFly}. These studies highlight frame selection and context compression as practical strategies for limiting the growth of historical input. However, these mechanisms focus primarily on \emph{what} historical information is retained and \emph{how much} of it is preserved, leaving the temporal boundary that determines \emph{when} an observation becomes available to the historical branch less explicitly characterized.

\subsection{Multi-Step Action Prediction and Generation}
\label{sec:related-action-diffusion}

Research on action chunking has examined not only how to jointly predict multiple future actions but also how to execute and update action chunks under closed-loop control. ACT uses a Transformer to predict an action sequence and applies temporal ensembling to aggregate overlapping predictions made at successive decision steps, thereby promoting temporally consistent continuous control \cite{zhaoLearningFineGrainedBimanual2023}. Bidirectional Decoding employs a test-time strategy designed to balance consistency within an action chunk against responsiveness to newly observed states, thereby supporting adaptive closed-loop execution \cite{liuBidirectionalDecodingImproving2024}. Real-Time Chunking addresses the deployment of diffusion- and flow-based policies by asynchronously generating the next action chunk while the current chunk is being executed, thereby reducing control delays caused by inference latency \cite{blackRealTimeExecutionAction2025}.

Diffusion-based policies provide a generative framework for multi-step action prediction. Diffusion Policy models continuous action trajectories through iterative denoising and deploys the predicted trajectories within a receding-horizon control loop with online replanning \cite{chiDiffusionPolicyVisuomotor2025}. In discrete action spaces, Dream-VL introduces a diffusion-based vision--language backbone with bidirectional attention, while Dream-VLA extends this backbone through continual pretraining on robotic data and jointly denoises multiple masked action tokens \cite{yeDreamVLDreamVLAOpen2026}.

Beyond direct action generation, another line of work augments planning with predicted future states or learned world dynamics. DreamVLA integrates compact world knowledge into multi-step action generation \cite{zhangDreamVLAVisionLanguageActionModel2025}, WorldVLN models latent world--action transitions \cite{WorldVLN}, and ImagineUAV predicts future visual observations to guide trajectory generation \cite{ImagineUAV}. Collectively, these approaches demonstrate the value of modeling both future actions and possible environmental evolution for multi-step planning. However, how discrete diffusion policies can exploit short-horizon structure among jointly predicted actions in aerial VLN, while executing only the current action and replanning from each new observation, remains comparatively underexplored.

\section{Method}
\label{sec:method}

\subsection{Problem Formulation}
\label{subsec:problem-formulation}

Given a natural-language instruction
$I=(w_1,w_2,\ldots,w_{N_I})$,
an aerial agent starts from an initial pose
$s_0=(p_0,\psi_0)$ in a 3D environment, where
$p_0=(x_0,y_0,z_0)$ denotes its position and $\psi_0$ denotes its heading.
At each decision step $t$, the agent receives an egocentric RGB observation
$O_t$ and selects an action $a_t$ from a discrete action space $\mathcal{A}$.
Executing $a_t$ moves the agent to a new pose $s_{t+1}$ and produces the next
observation $O_{t+1}$.

The action space consists of Stop, forward motion with different step sizes,
left and right turns, vertical movement, and lateral movement. The navigation
episode terminates when the agent selects Stop or reaches the maximum number of
decision steps. Given the ground-truth destination $p^*$, an episode is
considered successful if the Euclidean distance between the agent's final
position $p_T$ and the destination is less than $20$ m:
\[
\|p_T-p^*\|_2 < 20~\mathrm{m}.
\]

\subsection{Framework Overview}

\begin{figure}
	\centering
	\includegraphics[width=1.0\linewidth]{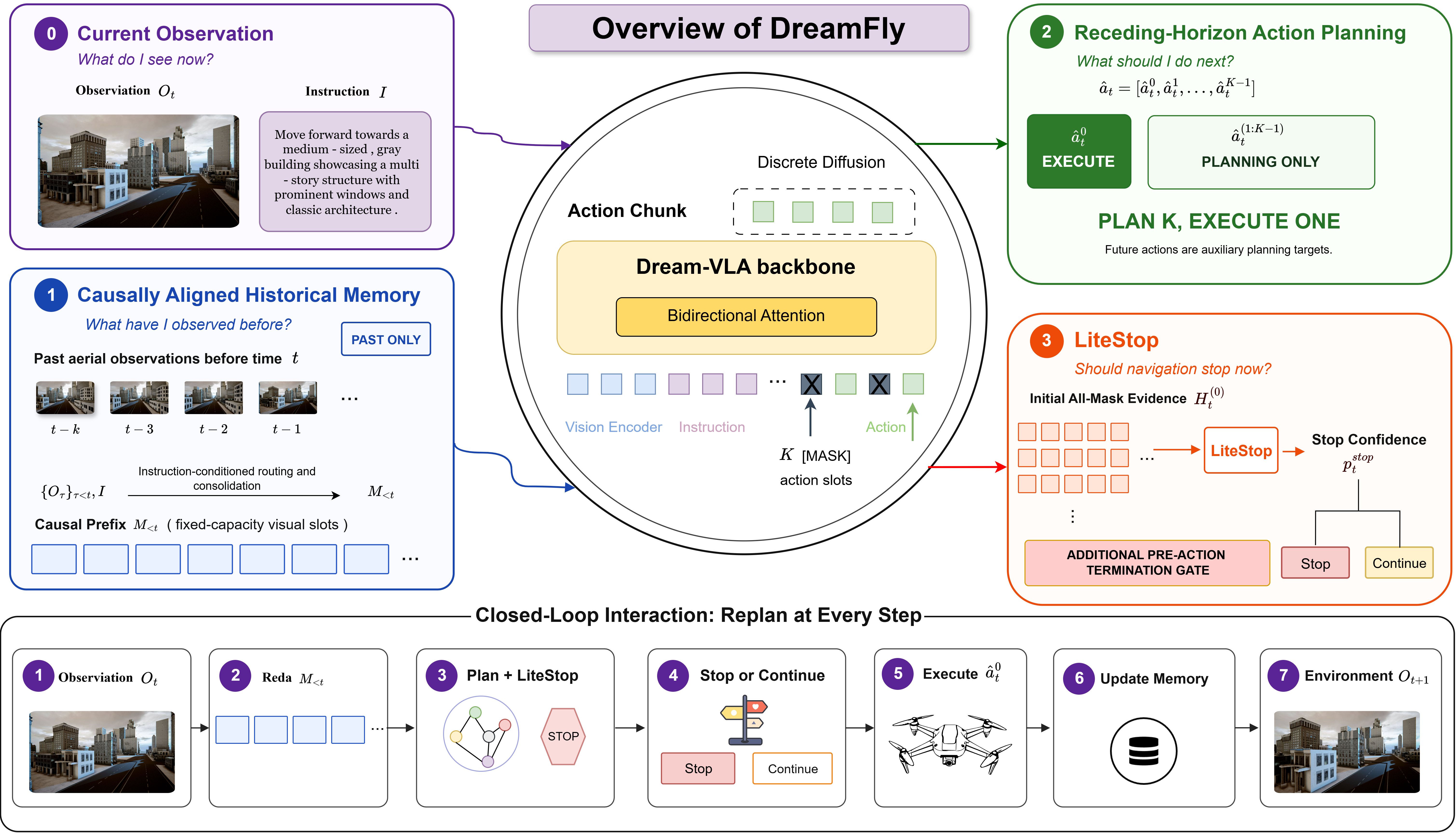}
	\caption{Overview of the proposed DreamFly framework. At each decision step, DreamFly conditions action prediction on the current observation, navigation instruction, and a causally aligned memory constructed from prior observations. The Dream-VLA backbone predicts a $K$-step action chunk through bidirectional diffusion, while LiteStop estimates termination from the initial all-mask action logits. If termination is not triggered, only the first action in the predicted chunk is executed before replanning from the next observation. Information derived from the current observation can enter the historical memory only for subsequent decisions.}
	\label{fig:framework}
\end{figure}

Fig.~\ref{fig:framework} illustrates the overall architecture of DreamFly, an aerial vision-language navigation framework built upon Dream-VLA~\cite{yeDreamVLDreamVLAOpen2026}. We adopt Dream-VLA as the policy backbone because its perception-to-action formulation aligns naturally with aerial VLN, directly mapping visual observations and language instructions to executable actions. Compared with autoregressive VLA backbones, Dream-VLA employs bidirectional diffusion modeling to support action chunking and parallel action prediction, providing a suitable foundation for receding-horizon planning.

Unlike the original VLA setting, aerial navigation requires retaining historical visual evidence and continually replanning as new observations arrive. DreamFly therefore introduces three key designs. First, a causally aligned historical memory $M_{<t}$ is constructed only from observations acquired before the current decision step and conditions the current visual representation, preserving a well-defined temporal information boundary while retaining previously observed landmarks. Second, DreamFly performs receding-horizon diffusion planning by jointly predicting a $K$-step action chunk. The predicted future actions provide short-horizon planning structure without being committed to consecutive execution. If termination is not triggered, only the first action is executed before the policy replans from the next observation. Third, LiteStop estimates termination from the initial all-mask action logits and provides a separate pre-action termination decision. Together, these components establish a temporally consistent closed-loop process that integrates historical reasoning, receding-horizon future-action planning, and explicit termination control.

\subsection{Causally Aligned Historical Memory}
\label{subsec:historical-memory}

During continuous aerial navigation, previously observed landmarks and scene cues may leave the current field of view while remaining relevant to subsequent decisions. Retaining all past observations would preserve such information, but would also introduce substantial redundancy and cause the visual input to grow with trajectory length. To maintain useful historical evidence under a bounded input budget, DreamFly compresses instruction-relevant information from past observations into a fixed-capacity historical memory. A short candidate FIFO and active tracks are maintained internally to accumulate evidence across recent observations, while only the long-term memory slots are exposed to the navigation policy.

At decision step $t$, let $O_{\tau}$ denote the observation acquired at  step ${\tau}$. The historical memory available to the policy is defined as
\[
M_{<t}
=
\mathcal{F}_{\mathrm{mem}}
\left(
I,
\left(O_{\tau}\right)_{\tau<t}
\right),
\]
where $\mathcal{F}_{\mathrm{mem}}$ denotes the fixed memory-construction operator. This definition constrains only the historical branch: the current observation $O_t$ remains directly available for action prediction, while information derived from $O_t$ can enter the historical memory only from subsequent decision steps.

\noindent\textbf{Instruction-conditioned candidate construction.}
For each observation, a frozen CLIPSeg dense router and a frozen OWLv2 region router extract complementary visual candidates conditioned on the complete navigation instruction $I$. To accommodate the finite text context of the routers without truncating the instruction, the full instruction is covered by fixed overlapping token windows. CLIPSeg provides dense instruction-relevant responses together with a visual feature grid, whereas OWLv2 proposes spatially localized regions. Rather than combining features from the two models, OWLv2 regions are represented directly in the frozen CLIPSeg visual feature space. For an OWLv2 region $b$, its visual representation is obtained as
\[
\mathbf{f}(b)
=
\operatorname{Norm}
\left(
\sum_g
\operatorname{area}
\left(
G_g\cap b
\right)
\mathbf{v}_g
\right),
\]
where $g$ indexes the CLIPSeg grid cells, $G_g$ denotes the spatial region of the $g$-th cell, $\mathbf{v}_g$ is its normalized visual feature, and $\operatorname{Norm}(\cdot)$ denotes $L_2$ normalization. CLIPSeg candidates directly use their corresponding grid features. Candidates are subsequently consolidated only when they exhibit sufficient spatial overlap and visual similarity.

\noindent\textbf{Evidence-driven long-term memory.}
The resulting candidates are associated with active tracks over recent observations according to visual and spatial consistency, allowing evidence for the same visual content to accumulate across observations. Candidates receiving repeated and stable support become eligible for \emph{persistent promotion}, whereas candidates without sufficient repeated support may enter memory through \emph{single-observation promotion} when they satisfy the confidence, region-validity, score-separation, and novelty criteria. These complementary mechanisms preserve both stable cross-observation evidence and informative content that may be visible only briefly.

Eligible candidates are ranked by their write utility, and at most two long-term write candidates are selected at each decision step. For each selected candidate, if a compatible slot is found, that slot is updated with the candidate; otherwise, the candidate is assigned to an empty slot or, when the memory is full, replaces an existing slot according to the retention criterion. 

Each valid slot maintains an \emph{anchor} and an optional \emph{prototype}. The anchor stores the visual feature of the concrete instance with the highest
historical write utility, whereas the prototype represents accumulated
compatible evidence once stable repeated support has been established. A candidate admitted through single-observation promotion may therefore form a valid anchor-only slot; such a slot is promoted to include a prototype when it is subsequently matched by a persistent write supported by stable cross-observation evidence.

At decision step $t$, the memory adapter represents the policy-facing memory as
\[
M_{<t}
=
\left\{
\left(
\mathbf{r}^{j}_{<t},
\mu^{j}_{<t}
\right)
\right\}_{j=1}^{16},
\qquad
\mathbf{r}^{j}_{<t}
=
\left[
\mathbf{e}^{j,\mathrm{anc}}_{<t};
\mathbf{e}^{j,\mathrm{pro}}_{<t};
\rho^{j}_{<t};
\log\left(1+\delta^{j}_{<t}\right)
\right],
\]
where $j$ indexes the memory slots, $\mu^{j}_{<t}\in\{0,1\}$ indicates slot validity, $\mathbf{e}^{j,\mathrm{anc}}_{<t}$ and $\mathbf{e}^{j,\mathrm{pro}}_{<t}$ denote the anchor and prototype features, respectively, $\rho^{j}_{<t}\in\{0,1\}$ indicates prototype presence, and $\delta^{j}_{<t}$ is the number of decision steps since the slot was last updated. An anchor-only slot remains valid; when no prototype is available, its prototype component is zero-filled and $\rho^{j}_{<t}=0$. For an invalid slot with $\mu^j_{<t}=0$, the complete input representation $\mathbf{r}^{j}_{<t}$ is zero-filled and is excluded from the key/value dimension
of cross-attention by the slot-validity mask.

\noindent\textbf{Memory-conditioned visual representation.}
DreamFly conditions the current visual representation on historical memory rather than appending memory slots as additional multimodal tokens. Let
\[
\mathbf{Z}_t
=
\operatorname{Projector}
\left(
\operatorname{VisionEncoder}(O_t)
\right)
\]
denote the current image tokens. Each slot representation $\mathbf{r}^{j}_{<t}$ is mapped by a memory slot encoder $\phi(\cdot)$, and the resulting slot embeddings are stacked in slot order:
\[
\mathbf{E}_{<t}
=
\begin{bmatrix}
	\phi(\mathbf{r}^{1}_{<t})\\
	\vdots\\
	\phi(\mathbf{r}^{16}_{<t})
\end{bmatrix}.
\]
Using the current image tokens as queries and the historical slot embeddings as keys and values, DreamFly retrieves historical context through
\[
\mathbf{C}_t
=
\operatorname{MHA}
\left(
\mathbf{Z}_t W_Q,
\mathbf{E}_{<t} W_K,
\mathbf{E}_{<t} W_V;
\boldsymbol{\mu}_{<t}
\right),
\]
where $W_Q$, $W_K$, and $W_V$ are learned projection matrices,
$\boldsymbol{\mu}_{<t}=(\mu^{1}_{<t},\ldots,\mu^{16}_{<t})$
is the slot-validity mask applied along the key/value slot dimension, and
$\mathbf{C}_t$ denotes the retrieved historical context. All valid slots participate in the masked cross-attention, allowing the current visual tokens to adaptively retrieve relevant historical evidence without hand-crafted top-$k$ selection.

The retrieved context is incorporated into the current visual tokens through a gated residual connection:
\[
\mathbf{G}_t
=
1+
\tanh
\left(
\mathbf{Z}_t W_G+\mathbf{b}_G
\right),
\qquad
\widetilde{\mathbf{Z}}_t
=
\mathbf{Z}_t
+
\mathbf{M}_{\mathrm{img}}
\odot
\mathbf{G}_t
\odot
\left(\mathbf{C}_t W_O\right),
\]
where $\mathbf{G}_t$ is the learned gate, $W_G$ and $\mathbf{b}_G$ are its parameters, $W_O$ is the residual output projection, $\mathbf{M}_{\mathrm{img}}$ is the valid-image-token mask, and $\odot$ denotes element-wise multiplication. The residual output projection $W_O$ and the gating parameters $W_G$ and
$\mathbf{b}_G$ are zero-initialized. Since $W_O=0$ at initialization, the
adapter initially implements the identity mapping
$\widetilde{\mathbf Z}_t=\mathbf Z_t$, while the initial gate value is
$\mathbf G_t=1$. The resulting memory-conditioned tokens $\widetilde{\mathbf{Z}}_t$ are then processed by the Dream-VLA backbone together with the navigation instruction for action prediction.

\noindent\textbf{Causal alignment across training and deployment.}
Training and closed-loop deployment share the same frozen memory-construction mechanism and temporal visibility boundary, but operate on different observation histories. Training uses prefix memories constructed along expert trajectories, whereas deployment initializes a fresh online memory for each rollout and updates it from observations actually acquired by the agent. Consequently, the resulting memory states may differ, while the causal constraint remains unchanged: the memory $M_{<t}$ used at step $t$ contains only information acquired before that decision step.

\subsection{Receding-Horizon Diffusion Planning}
\label{subsec:action-planning}

Predicting only the next action provides an immediate control decision but
does not explicitly represent the short-horizon action structure surrounding
that decision. DreamFly therefore extends single-step prediction to a
fixed-horizon discrete action chunk and exploits the bidirectional Dream-VLA
backbone to jointly model the current action together with $K-1$ near-future
actions. The predicted chunk serves as a structured short-horizon planning
representation rather than an open-loop command sequence. At each decision
step, DreamFly generates the complete chunk. LiteStop may terminate the episode
before action execution; otherwise, at most the leading action is handled, and
the policy replans after a non-terminal motion produces new visual feedback.

\noindent\textbf{Discrete action-chunk formulation.}
Given the memory-conditioned visual representation
$\widetilde{\mathbf{Z}}_t$ and the navigation instruction $I$, DreamFly
predicts a length-$K$ discrete action chunk
\[
\hat{\mathbf{a}}_t
=
\left[
\hat{a}_t^0,
\hat{a}_t^1,
\ldots,
\hat{a}_t^{K-1}
\right],
\qquad
\hat{a}_t^h\in\mathcal{A},
\]
where $\mathcal{A}$ is the discrete aerial action space and $K$ is the
planning horizon. The first slot $\hat{a}_t^0$ corresponds to the current
control decision, whereas the remaining slots $\hat{a}_t^{1:K-1}$ represent
short-horizon future actions predicted under the same current context. These
future predictions form part of the structured short-horizon plan but are not
committed commands to be executed consecutively.

Let $\mathcal{V}$ denote the full Dream-VLA vocabulary, and let
\[
\chi:\mathcal{A}\hookrightarrow\mathcal{V}
\]
denote the injective mapping from each discrete action to its dedicated action
token. Its image $\chi(\mathcal{A})\subset\mathcal{V}$ forms the dedicated
action-token set. This distinction allows us to use environment-level actions
in the planning formulation while explicitly identifying their corresponding
vocabulary targets during model training and diffusion generation.

\noindent\textbf{Joint masked action learning.}
During training, the action chunk is supervised by the suffix beginning at
decision step $t$ of the stored action sequence. Let
$(a_0^\star,\ldots,a_{T-1}^\star)$ denote the length-$T$ action sequence
stored for a training trajectory. At a decision step
$t\in\{0,\ldots,T-1\}$, the number of available actions within the prediction
horizon is
\[
L_t
=
\min(K,T-t),
\qquad
v_{t,h}
=
\mathbb{I}[h<L_t],
\]
where $v_{t,h}\in\{0,1\}$ indicates whether the $h$-th action slot belongs to
the available supervision prefix. To preserve a fixed-length template near
the end of a trajectory, positions beyond this prefix are padded with the Stop
action. We denote the padded target by
\[
\bar{a}_{t,h}^{\star}
=
\begin{cases}
	a_{t+h}^{\star}, & h<L_t,\\[1mm]
	a_{\mathrm{stop}}, & h\ge L_t,
\end{cases}
\]
where $a_{\mathrm{stop}}\in\mathcal{A}$ denotes the Stop action. The validity
mask ensures that Stop tokens appended beyond the length-$T$ supervision
sequence solely to complete the fixed $K$-slot template contribute no training
loss. Any Stop token already contained in
$\mathbf{a}^\star=(a_0^\star,\ldots,a_{T-1}^\star)$ remains a valid supervised
target.

Importantly, extending the prediction horizon does not expose future visual
observations to the policy. At decision step $t$, the model is conditioned on
the current visual representation $\widetilde{\mathbf{Z}}_t$ and instruction
$I$; future observations $O_{t+1},\ldots,O_{t+K-1}$ are not provided.
Supervision for future action slots is provided only by the corresponding
action targets.

During training, all $K$ action positions supplied to the bidirectional
backbone are replaced with \texttt{[MASK]}. Only the valid action prefix is
used for supervision, while synthetic tail padding is excluded by the
valid-prefix mask. DreamFly then jointly predicts all action slots in a single
bidirectional forward pass. Thus, action-policy training does not unroll
iterative denoising; during inference, iterative discrete diffusion
progressively resolves the masked action slots.

At decision step $t$, let
\[
\mathbf{z}_{t,h}
\in
\mathbb{R}^{|\mathcal{V}|},
\qquad h=0,\ldots,K-1,
\]
denote the shifted full-vocabulary logit vector at the sequence position that
predicts the target token $\chi\!\left(\bar{a}_{t,h}^{\star}\right)$ for the
$h$-th action slot.

Let $\mathcal{U}_t$ denote the valid non-action and non-padding positions in
the shifted sequence, and let $q_{t,h}$ denote the shifted sequence position
whose logits form $\mathbf{z}_{t,h}$. We define the deterministic geometric
kernel
\[
\kappa_{ij}
=
\begin{cases}
	0, & i=j,\\[1mm]
	\dfrac{\beta_{\mathrm{car}}}{2}
	\left(1-\beta_{\mathrm{car}}\right)^{|i-j|-1},
	& i\neq j,
\end{cases}
\]
and compute the context coefficient as
\[
c_{t,h}
=
\max
\left(
10^{-6},
\sum_{i\in\mathcal{U}_t}
\kappa_{i,q_{t,h}}
\right),
\qquad
\beta_{\mathrm{car}}=0.1.
\]
All action slots, including unsupervised tail slots, are excluded from
$\mathcal{U}_t$. Let $\mathcal{B}$ denote a minibatch of decision-step
samples, with trajectory identities omitted for clarity. The action objective
is
\[
\mathcal{L}_{\mathrm{act}}
=
\frac{
	\displaystyle
	\sum_{t\in\mathcal{B}}
	\sum_{h=0}^{K-1}
	v_{t,h}\,
	c_{t,h}\,
	\gamma^h\,
	\operatorname{CE}_{\mathcal{V}}
	\left(
	\mathbf{z}_{t,h},
	\chi\!\left(\bar{a}_{t,h}^{\star}\right)
	\right)
}{
	\displaystyle
	\sum_{t\in\mathcal{B}}
	\sum_{h=0}^{K-1}
	v_{t,h}\,
	c_{t,h}\,
	\gamma^h
},
\]
where $0<\gamma\leq1$ controls horizon-dependent weighting; values below one
progressively emphasize near-term action targets. The validity mask $v_{t,h}$
restricts the objective to positions contained in the length-$T$ supervision
sequence. Following Dream-VLA, the cross-entropy is evaluated over the complete
vocabulary $\mathcal{V}$. The restriction to the dedicated action-token set
$\chi(\mathcal{A})$ is applied only during diffusion generation to guarantee
executable action outputs.

\noindent\textbf{Discrete diffusion generation.}
At inference time, DreamFly generates the action chunk through iterative
discrete diffusion. Let $\mathbf{m}_t^{(s)}$ denote the abstract state of the
$K$ action slots at denoising step $s$; fixed formatting tokens in the
assistant scaffold are omitted from this notation. Generation begins from
\[
\mathbf{m}_t^{(0)}
=
\left[
\texttt{[MASK]},
\ldots,
\texttt{[MASK]}
\right],
\]
and proceeds through
\[
\mathbf{m}_t^{(0)}
\rightarrow
\mathbf{m}_t^{(1)}
\rightarrow
\cdots
\rightarrow
\mathbf{m}_t^{(S)},
\qquad
\Pi_{\mathcal{A}}
\left(
\mathbf{m}_t^{(S)}
\right)
=
\hat{\mathbf{a}}_t,
\]
where $S$ denotes the number of diffusion steps and
$\Pi_{\mathcal{A}}(\cdot)$ decodes the $K$ action slots through
$\chi^{-1}$. This notation distinguishes the predicted action chunk from the
complete assistant scaffold, which also contains fixed formatting tokens.

We use the monotonic \emph{origin} sampler inherited from Dream-VLA. Let
$\{\xi_s\}_{s=0}^{S}$ be linearly spaced from $1$ to
$\varepsilon_{\mathrm{diff}}=10^{-3}$. At denoising step
$s\in\{0,\ldots,S-1\}$, each unresolved action slot is independently
transferred from \texttt{[MASK]} to a sampled action token with probability
\[
\omega_s
=
\begin{cases}
	1-\dfrac{\xi_{s+1}}{\xi_s}, & s<S-1,\\[2mm]
	1, & s=S-1.
\end{cases}
\]
Conditional on transfer, the replacement token is sampled from the model
distribution after suppressing all logits outside $\chi(\mathcal{A})$.
Resolved action slots are not remasked, and the final denoising step resolves
all remaining masks.

Unlike fixed left-to-right autoregressive decoding, the bidirectional backbone
allows unresolved action slots to share the available context, and multiple
slots may be resolved within the same denoising step. Resolved action tokens
can subsequently provide context for slots that remain unresolved. Before
sampling at each unresolved position, logits outside $\chi(\mathcal{A})$ are
suppressed; consequently, every transferred token belongs to the dedicated
action-token set.

The initial all-mask forward also provides the planning representation consumed
by LiteStop. Let
$\mathbf{z}_{t,h}^{(0)}\in\mathbb{R}^{|\mathcal{V}|}$ denote the shifted
full-vocabulary logit vector aligned with the $h$-th action position at the
initial denoising step. We retain the corresponding action-token slices as
\[
\mathbf{H}_t^{(0)}
=
\left[
\operatorname{Slice}_{\chi(\mathcal{A})}
\left(\mathbf{z}_{t,0}^{(0)}\right);
\ldots;
\operatorname{Slice}_{\chi(\mathcal{A})}
\left(\mathbf{z}_{t,K-1}^{(0)}\right)
\right]
\in
\mathbb{R}^{K\times|\mathcal{A}|}.
\]
Here, $\operatorname{Slice}_{\chi(\mathcal{A})}$ extracts the action-token
coordinates in the fixed action-ID order. The representation is extracted
during the initial denoising forward, while all $K$ action positions remain
masked and before any action token is transferred. It therefore records the
policy responses at all $K$ action positions under the same all-mask planning
state. Importantly, retaining $\mathbf{H}_t^{(0)}$ does not terminate or
shorten diffusion generation; the complete action chunk is generated before
LiteStop is evaluated.

\noindent\textbf{Receding-horizon execution.}
After the complete action chunk has been generated, LiteStop determines whether
navigation should terminate before any chunk action is executed. If LiteStop
is triggered, the episode terminates without executing an action from
$\hat{\mathbf{a}}_t$. Otherwise, the leading action $\hat{a}_t^0$ is handled
according to the original action-level semantics. If
$\hat{a}_t^0=a_{\mathrm{stop}}$, the episode terminates through the action-level
Stop condition; otherwise, the selected motion is executed. The remaining
predictions $\hat{a}_t^1,\ldots,\hat{a}_t^{K-1}$ are discarded rather than
cached for subsequent execution.

After a non-terminal motion is executed, the agent receives the next
observation $O_{t+1}$ and replans a complete action chunk. The historical
memory available at the next decision step follows the same causal definition
introduced in Sec.~\ref{subsec:historical-memory}:
\[
M_{<t+1}
=
\mathcal{F}_{\mathrm{mem}}
\left(
I,
\left\{O_{\tau}\right\}_{\tau\leq t}
\right).
\]
Thus, information derived from $O_t$ may influence the historical context at
step $t+1$ only if it is retained by the memory-construction mechanism,
whereas the newly acquired observation $O_{t+1}$ remains a separate current
visual input and is not simultaneously exposed through the historical-memory
branch. DreamFly therefore follows a receding-horizon strategy that plans $K$
actions at each decision step and executes at most the leading action before
replanning from newly observed visual evidence.

\subsection{LiteStop: Decoupled Termination Control}
\label{subsec:litestop}

Termination has a qualitatively different consequence from ordinary motion decisions: an erroneous motion may be corrected through subsequent replanning, whereas premature termination immediately ends the episode. We therefore introduce \emph{LiteStop}, a lightweight auxiliary head that derives an additional decision-level termination signal from the frozen policy's initial all-mask planning response. LiteStop is optimized separately from the action policy: it consumes the frozen policy's planning logits, but does not update policy parameters or modify the original action-level Stop semantics.

\noindent\textbf{All-mask termination representation.}
We reuse the initial planning representation
$\mathbf H_t^{(0)}\in\mathbb R^{K\times|\mathcal A|}$ defined in
Sec.~\ref{subsec:action-planning}. It contains the shift-aligned raw logits
over the dedicated action-token set $\chi(\mathcal A)$ for all $K$ planning
positions. LiteStop maps this complete response grid to a scalar Stop logit:
\[
\ell_t^{\mathrm{stop}}
=
g_{\mathrm{stop}}\!\left(\mathbf H_t^{(0)}\right)
=
W_2\operatorname{SiLU}\!\left(
W_1\operatorname{LN}\!\left(
\operatorname{vec}\!\left(\mathbf H_t^{(0)}\right)
\right)
+\mathbf b_1
\right)
+b_2,
\qquad
p_t^{\mathrm{stop}}
=
\sigma\!\left(\ell_t^{\mathrm{stop}}\right).
\]
Here, $\operatorname{vec}(\cdot)$ vectorizes the logit grid,
$\operatorname{LN}(\cdot)$ denotes layer normalization, and
$g_{\mathrm{stop}}$ contains the learnable LayerNorm and MLP parameters.
Using the complete $K\times|\mathcal A|$ grid allows LiteStop to exploit the
policy's short-horizon planning state rather than relying only on the
first-position Stop logit.

\noindent\textbf{Frozen-policy supervision.}
Let $a_t^\star$ denote the expert action at decision step $t$. The binary Stop target is
\[
y_t^{\mathrm{stop}}
=
\mathbb{I}\!\left[
a_t^\star=a_{\mathrm{stop}}
\right].
\]
A Stop appearing only at a future chunk position or in synthetic tail padding
therefore does not affect the current label. The supervision uses neither
geometric success nor terminal metadata, and thus calibrates the frozen
policy's action-level Stop tendency rather than learning an independent
goal-reached classifier.

For a training batch $\mathcal B$, we minimize
\[
\mathcal L_{\mathrm{stop}}
=
-\frac{1}{|\mathcal B|}
\sum_{t\in\mathcal B}
\left[
\lambda_+ y_t^{\mathrm{stop}}\log p_t^{\mathrm{stop}}
+
\left(1-y_t^{\mathrm{stop}}\right)
\log\!\left(1-p_t^{\mathrm{stop}}\right)
\right],
\]
where $\lambda_+=4.0$ is the fixed positive-class weight. The complete navigation
policy, including its visual, memory, and action-planning components, remains
frozen, and only LiteStop is optimized. During training,
$\mathbf H_t^{(0)}$ is extracted by a single all-mask bidirectional forward
pass without unfolding iterative diffusion, matching the all-mask
representation semantics used at deployment.

\noindent\textbf{Pre-action termination.}
During inference, $\mathbf H_t^{(0)}$ is cached from the initial all-mask
denoising forward. The frozen policy nevertheless completes the full diffusion
process and obtains the complete decoded action chunk before LiteStop is
evaluated. LiteStop therefore requires no additional backbone forward pass,
but it is not a diffusion early-exit mechanism. This ordering preserves the
frozen policy's complete generation path and confines LiteStop to pre-action
termination control; computational early exit is outside the scope of the
present design.

Given a fixed threshold $\eta_{\mathrm{stop}}$, the LiteStop decision is
\[
d_t^{\mathrm{stop}}
=
\mathbb{I}\!\left[
p_t^{\mathrm{stop}}\ge\eta_{\mathrm{stop}}
\right].
\]
The final termination decision combines LiteStop with the frozen policy's
action-level Stop condition:
\[
d_t^{\mathrm{term}}
=
d_t^{\mathrm{stop}}
\lor
\mathbb{I}\!\left[
\hat a_t^0=a_{\mathrm{stop}}
\right].
\]
If $d_t^{\mathrm{stop}}=1$, the episode terminates before any action from the
current chunk is executed. Otherwise, the first action $\hat a_t^0$ is handled
under the receding-horizon protocol from
Sec.~\ref{subsec:action-planning}. If
$\hat a_t^0=a_{\mathrm{stop}}$, the episode terminates through the retained
action-level Stop condition; if neither termination condition is satisfied,
the selected motion is executed and the agent acquires a new observation
before replanning the complete action chunk. LiteStop therefore provides an
additional pre-action termination pathway rather than vetoing the frozen
policy's action-level Stop condition.

\noindent\textbf{Overall closed-loop operation.}
Taken together, the three components form a causal receding-horizon control
loop. At decision step $t$, DreamFly combines the current observation $O_t$
with historical memory $M_{<t}$ to obtain the memory-enhanced visual
representation $\widetilde{\mathbf Z}_t$. Together with the navigation
instruction $I$, this representation conditions the action policy, which
generates a complete $K$-step action chunk $\hat{\mathbf a}_t$ while retaining
the initial all-mask planning representation $\mathbf H_t^{(0)}$ for LiteStop.
After the complete action chunk has been generated and before any action from
the chunk is executed, LiteStop determines whether the episode should
terminate. If termination is not triggered, the leading action $\hat a_t^0$ is
handled according to the original action-level semantics.

The historical memory available to the current decision contains only
information derived from observations preceding $O_t$. Information extracted
from $O_t$ may affect the historical memory only from subsequent decisions
onward and can therefore first become accessible through $M_{<t+1}$ if it is
retained by the memory-construction mechanism. After a non-terminal motion
produces a new observation $O_{t+1}$, DreamFly repeats the same process using
$O_{t+1}$ as the current visual input together with the corresponding causal
memory prefix.

The components are optimized in stages rather than through a single joint
objective. Historical memory construction remains fixed while the
memory-conditioned action policy is trained with the horizon-aware action
objective $\mathcal L_{\mathrm{act}}$. After the navigation policy has been
trained, the complete policy is frozen and only LiteStop is optimized with
$\mathcal L_{\mathrm{stop}}$. Consequently, LiteStop training does not
back-propagate into the visual, memory-adaptation, or action-planning
components. This staged design integrates causal historical context,
short-horizon diffusion planning, and decoupled termination calibration
within a unified closed-loop navigation policy.

\section{Experiments}
\label{sec:experiments}

\subsection{Experimental Setup}
\label{sec:experimental-setup}

\subsubsection{Datasets}
\label{sec:datasets}

We conduct our experiments based on the OpenFly dataset. Before training, we apply four standardization and augmentation steps to the released data. First, we correct the 8-D action vector associated with \emph{Forward 6m} to its canonical encoding. Second, approximately 190,000 decision steps with non-standard action labels are remapped to the corresponding canonical actions, with $-1$ mapped to \emph{Go Up} and $-2$ to \emph{Go Down}. Third, we remove the pre-packaged historical keyframes and retain only the current RGB observation at each decision step. Fourth, We additionally store$[\Delta x,\Delta y,\Delta z,\mathrm{yaw}]$ as provenance metadata. This metadata is not used as an input to either historical-memory construction or policy inference. These relative poses are not used as additional policy inputs during closed-loop inference. The resulting training set contains 20 subsets with 85,785 trajectories and 1,356,622 decision steps.

\begin{figure}[htbp]
	\centering
	\begin{overpic}[width=0.8\linewidth]{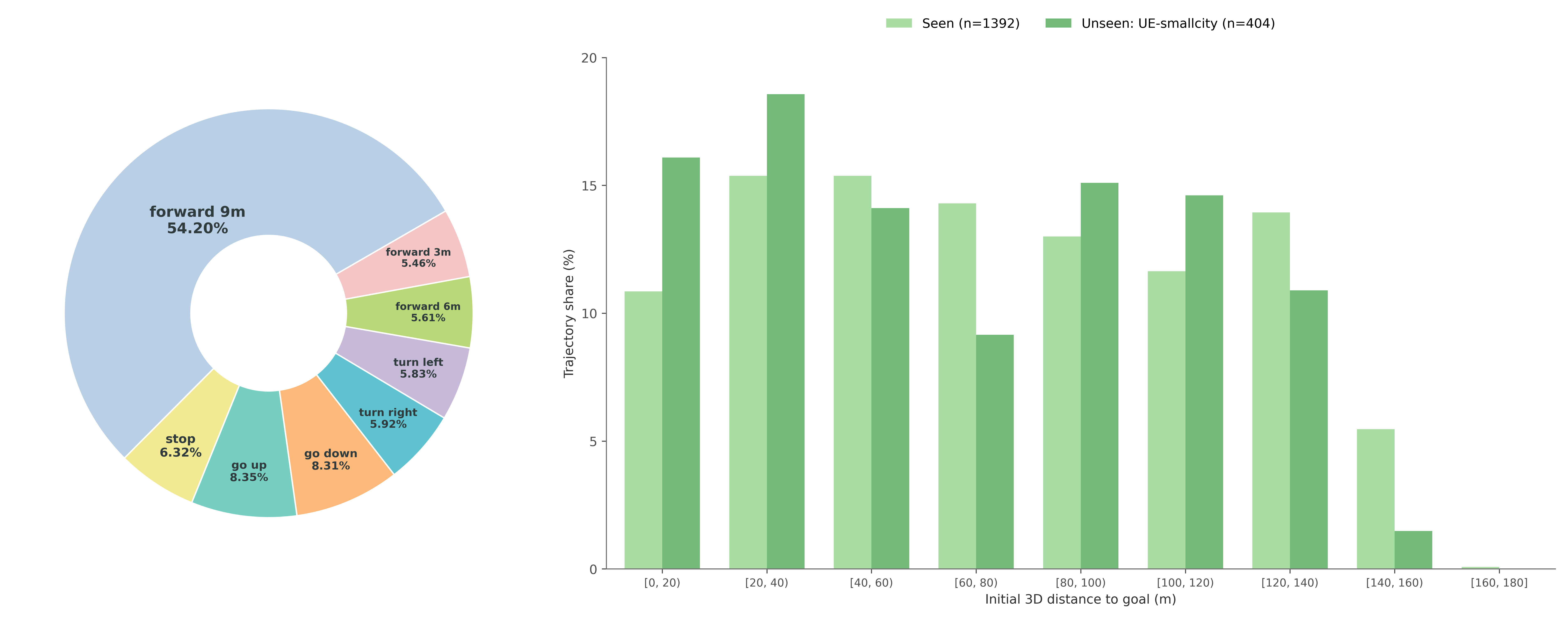}
		\put(15,-3){\large\textbf{(a)}}
		\put(62,-3){\large\textbf{(b)}}
	\end{overpic}
	\caption{Dataset statistics:
		(a) action distribution in the training set;
		(b) distributions of initial goal distances on the test-seen and test-unseen splits.}
	\label{fig:dataset}
\end{figure}

Our evaluation covers eight AirSim/UE environments with 1,796 trajectories. The test-seen split contains UE BigCity and six AirSim urban environments, totaling 1,392 trajectories, whereas the test-unseen split contains 404 trajectories from UE SmallCity. This evaluation therefore covers both Unreal Engine and AirSim environments and assesses navigation performance in seen environments as well as generalization to an unseen urban scene. As shown in Fig.~\ref{fig:dataset}, the training data exhibit a pronounced imbalance toward forward actions, while the test-seen and test-unseen splits show different distributions of initial goal distances.

\subsubsection{\texorpdfstring{\textbf{Evaluation Metrics.}}{Evaluation Metrics.}}
\label{evaluation-metrics.}

Following the evaluation protocol of OpenFly and prior aerial VLN studies, we adopt four standard metrics: navigation error (NE), success rate (SR), oracle success rate (OSR), and success weighted by path length (SPL). NE measures the average Euclidean distance between the UAV's final stopping position and the ground-truth target position, with a lower value indicating more accurate goal localization. SR measures the proportion of successful trajectories, where a trajectory is considered successful if the UAV stops within 20\,m of the target. OSR measures the proportion of trajectories that come within 20\,m of the target at any point during navigation, regardless of the final stopping position, and therefore reflects whether the agent successfully reaches the target vicinity. SPL jointly evaluates navigation success and path efficiency by weighting each successful trajectory according to the ratio between the shortest-path distance and the actual traveled path length. Higher values of SR, OSR, and SPL indicate better performance.

\subsubsection{Implementation Details}
\label{sec:implementation-details}

The Dream-VLA backbone is fine-tuned with all-linear LoRA
($r=32$, $\alpha=16$), while the memory-fusion adapter is trained jointly
and the base projector remains frozen. We use an action-chunk length of
$K=4$, horizon decay $\gamma=0.7$, and CAR reweighting probability $p=0.1$.
Training uses AdamW with a learning rate of $1\times10^{-4}$ and batch size 8
for up to 10,000 optimization steps. The navigation checkpoint at step 5,000
is used to train LiteStop and for closed-loop evaluation. Inference uses
12 discrete diffusion steps.

The historical memory contains 16 long-term slots with 512-dimensional
features. Memory is fused with the current visual tokens using a gated
cross-attention module with dimension 512 and eight attention heads.

\noindent\textbf{LiteStop operating-point selection.}
We evaluate $\eta_{\mathrm{stop}}\in{0.50,0.65,0.80}$ using the
step-500 LiteStop checkpoint on a balanced 64-trajectory calibration set
(8 per environment), disjoint from the final evaluation split.
Based on the overall trade-off in Table~\ref{tab:litestop-threshold},
We select $\eta_{\mathrm{stop}}=0.50$ because it yields the smallest OSR--SR gap and a slightly lower NE than $\eta_{\mathrm{stop}}=0.80$ while preserving the same SR. Unlike $\eta_{\mathrm{stop}}=0.80$, it also produces nontrivial LiteStop interventions.

\begin{table}[t]
	\centering
	\caption{Pilot selection of the LiteStop termination threshold.
		$N_{\mathrm{LS}}$ is the number of episodes terminated by LiteStop itself,
		out of the 64 pilot episodes. The selected operating point is marked by
		$\dagger$.}
	\label{tab:litestop-threshold}
	\setlength{\tabcolsep}{4pt}
	\begin{tabular}{cccccc}
		\toprule
		$\eta_{\mathrm{stop}}$
		& SR $\uparrow$
		& SPL $\uparrow$
		& OSR--SR $\downarrow$
		& NE $\downarrow$
		& $N_{\mathrm{LS}}$ \\
		\midrule
		$0.50^{\dagger}$ & 0.266 & 0.229 & 0.188 & 53.07 & 45/64 \\
		$0.65$           & 0.250 & 0.202 & 0.250 & 51.42 & 19/64 \\
		$0.80$           & 0.266 & 0.230 & 0.219 & 53.20 &  0/64 \\
		\bottomrule
	\end{tabular}
\end{table}

\subsection{Experiment Result}
\label{Experiment-Result}

As shown in Table~\ref{tab:comparison}, we compare DreamFly with six baselines: Random, Action Sampling, Seq2Seq, CMA, AerialVLN, and OpenFly-Agent. Random uniformly samples an action from the discrete action space at each step and executes it until either $\mathrm{Stop}_{\mathrm{tok}}$ is sampled or the maximum number of navigation steps is reached. Action Sampling follows the same procedure but samples actions according to their empirical distribution in the training set, providing a stronger stochastic baseline that reflects the action prior of the dataset. For OpenFly-Agent, we directly evaluate the officially released checkpoint in the same test environment to avoid introducing discrepancies by reproducing its original training and historical-observation pipeline under our standardized data protocol. All other learning-based baselines are trained on our processed training data for the same number of optimization steps.

\begin{figure}
	\centering
	\includegraphics[width=0.7\linewidth]{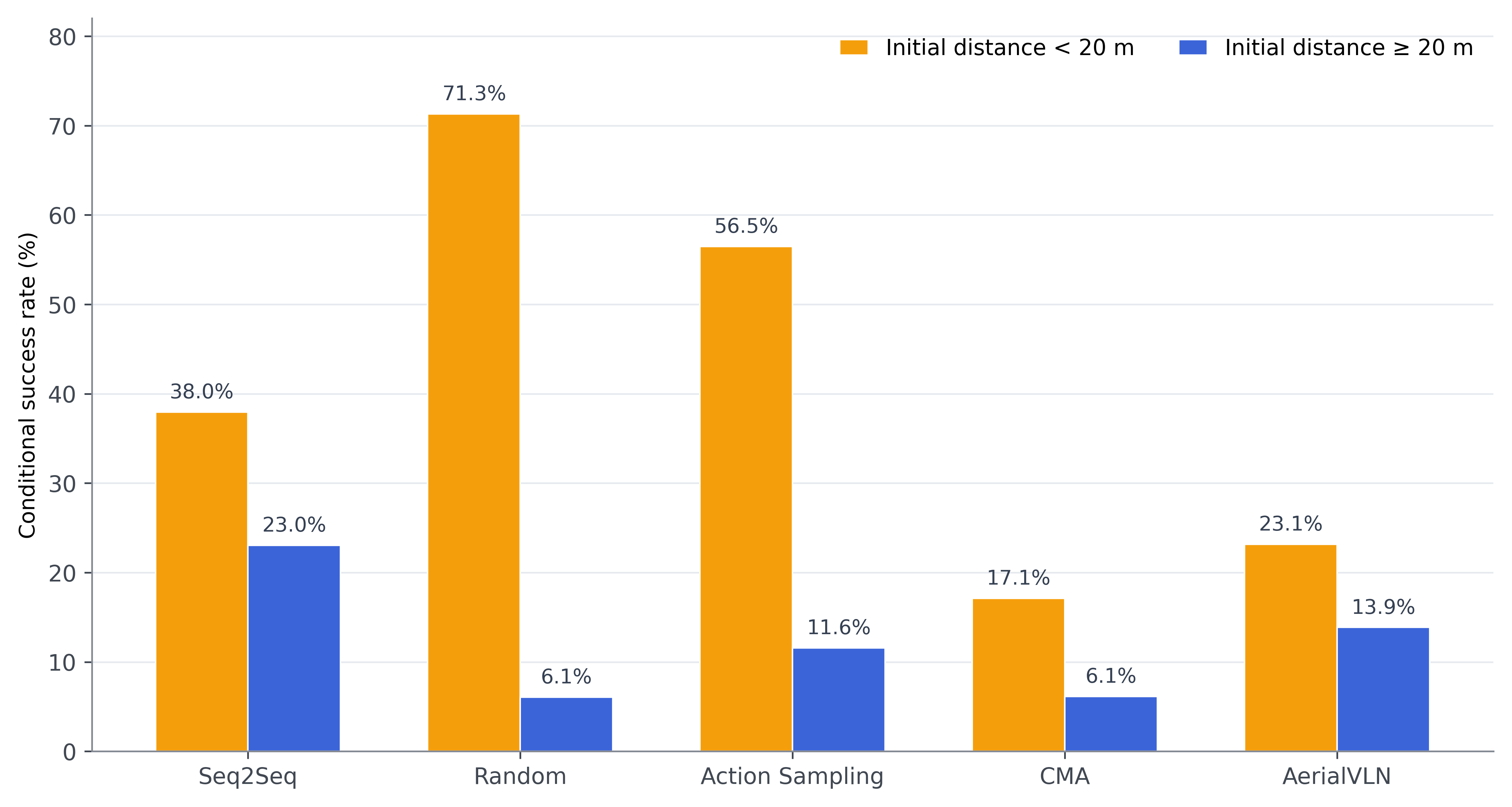}
	\caption{Conditional success rates of baseline methods under different initial goal distances. Trajectories are partitioned according to whether the initial goal distance is within the $20\,\mathrm{m}$ success radius. The stochastic baselines exhibit substantially higher success rates when initialized inside the success region, highlighting the strong influence of favorable initial configurations on their non-zero overall success rates.}
	\label{fig:baselinesucc}
\end{figure}

As shown in Figure~\ref{fig:baselinesucc}, the non-zero success rates of the stochastic baselines should not be interpreted as evidence of effective goal-directed navigation. To examine the influence of the initial state, we further partition the test trajectories according to whether their initial goal distance is within the $20\,\mathrm{m}$ success radius and report the conditional success rates for the two groups. Both Random and Action Sampling exhibit substantially higher success rates when the agent is initialized within the success radius than when it starts outside this region. This pronounced gap indicates that their non-zero overall success rates are strongly influenced by favorable initial configurations, rather than reflecting navigation policies that consistently drive the agent toward the target.

\begin{table}[t]
	\centering
	\caption{Performance comparison on the seen and unseen splits.}
	\label{tab:comparison}
	\resizebox{\linewidth}{!}{
		\begin{tabular}{lcccccccc}
			\toprule
			Method &
			\multicolumn{4}{c}{test-seen} &
			\multicolumn{4}{c}{test-unseen} \\
			\cmidrule(lr){2-5}
			\cmidrule(lr){6-9}
			& NE$\downarrow$ & SR$\uparrow$ & OSR$\uparrow$ & SPL$\uparrow$
			& NE$\downarrow$ & SR$\uparrow$ & OSR$\uparrow$ & SPL$\uparrow$ \\
			\midrule

			Random
			& 65.67m & 13.51\% & 18.75\% & 9.72\%
			& 59.99m & 15.35\% & 23.76\% & 11.31\% \\
			
			Action Sampling
			& 62.78m & 15.95\% & 26.51\% & 13.67\%
			& 55.27m & 20.54\% & 32.67\% & 17.22\% \\
			
			Seq2Seq
			& \underline{54.44m} & \underline{24.35\%} & 61.93\% & 19.35\%
			& \underline{47.69m} & \underline{26.49\%} & 61.88\% & \underline{19.62\%} \\
			
			CMA
			& 313.03m & 7.97\% & \textbf{69.32\%} & 6.26\%
			& 230.05m & 5.69\% & \textbf{73.02\%} & 3.92\% \\
			
			AerialVLN
			& 176.29m & 16.52\% & \underline{65.66\%} & 14.63\%
			& 161.19m & 9.65\% & \underline{68.07\%} & 7.93\% \\
			
			OpenFly-Agent
			& 122.89m & 22.63\% & 52.73\% & \underline{20.42\%}
			& 163.87m & 14.11\% & 62.38\% & 12.49\% \\
			
			DreamFly(Ours)
			& \textbf{44.87m} & \textbf{32.04\%} & 46.77\% & \textbf{28.22\%}
			& \textbf{45.29m} & \textbf{29.46\%} & 46.78\% & \textbf{23.54\%} \\

			\bottomrule
		\end{tabular}
	}
\end{table}

\subsection{Ablation Studies}\label{ablation-studies}

To evaluate the contribution of each component, we conduct both progressive and leave-one-out ablation studies, as reported in Table~\ref{tab:ablation}. Starting from the Dream-VLA baseline, introducing the causally aligned historical memory consistently improves navigation performance, while receding-horizon diffusion planning further improves the results by jointly modeling the current action and short-horizon future actions. Incorporating LiteStop yields the strongest overall performance. Conversely, removing Memory, Action Chunk, or LiteStop from the complete framework results in performance degradation. These consistent trends indicate that historical context modeling, future-action planning, and explicit termination provide complementary benefits to DreamFly.

\begin{figure}
	\centering
	\includegraphics[width=1.0\linewidth]{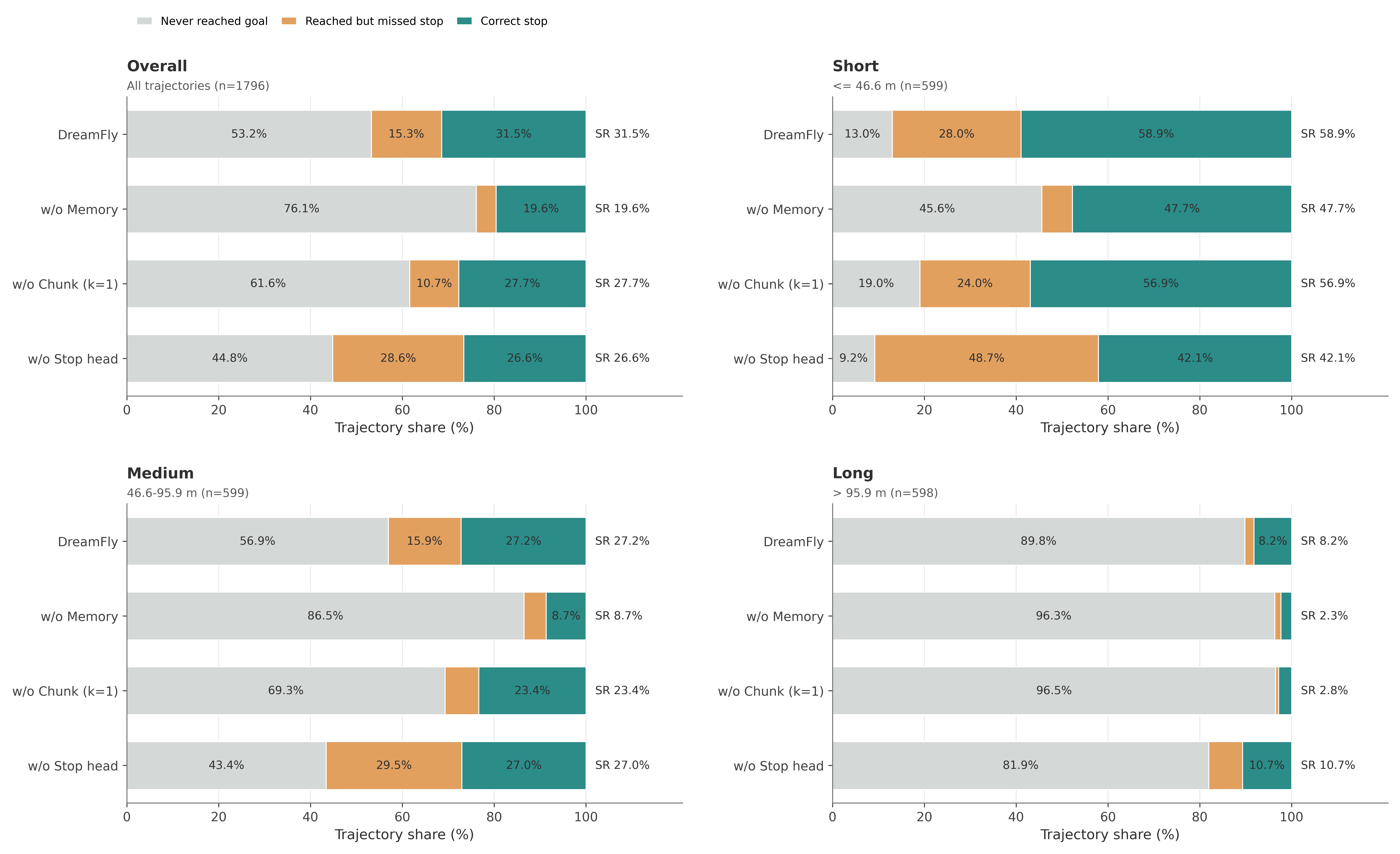}
	\caption{Component-wise analysis across different initial 3D Euclidean distance(s) to the goal. Test trajectories are partitioned into three groups according to their initial shortest-path distances to examine how the contribution of each DreamFly component varies with navigation distance.}
	\label{fig:analysis-of-temporal-decision}
\end{figure}

To further examine how these components contribute under different navigation distances, we partition the test trajectories into three groups according to their initial shortest-path distance. As shown in Fig.~\ref{fig:analysis-of-temporal-decision}, LiteStop provides its largest gain in the shortest-distance group, where successful navigation depends more directly on recognizing the appropriate termination point. Historical memory contributes more prominently at intermediate distances, while for larger initial distances, memory and action-chunk planning provide complementary benefits by maintaining cross-step visual context and introducing short-horizon future-action structure into each replanning step.

The contribution of LiteStop decreases as the initial distance increases, which is consistent with termination becoming relevant only after the agent reaches the vicinity of the goal. In this regard, the gap between OSR and SR provides a complementary view of termination behavior, since OSR reflects whether the trajectory reaches the success region whereas SR additionally requires successful termination within it. Overall, the distance-wise results further reveal the distinct roles of historical memory, receding-horizon diffusion planning, and LiteStop in the DreamFly framework.

\begin{table}[t]
	\centering
	\caption{Ablation study of different experimental settings.}
	\label{tab:ablation}
	\begin{tabular}{lcccc}
		\toprule
		Experiment & NE$\downarrow$ & SR$\uparrow$ & OSR$\uparrow$ & SPL$\uparrow$ \\
		\midrule
		Dream-VLA (baseline)& 67.82m & 21.55\% & 42.32\% & 16.09\% \\
		+ Causal Memory     & 48.93m & 24.11\% & 48.22\% & 19.85\% \\
		w/o Memory		 	& 62.02m & 19.60\% & 23.89\% & 18.76\% \\
		w/o Chunk		 	& 46.72m & 27.73\% & 38.42\% & 23.77\% \\
		w/o LiteStop		& 50.69m & 26.61\% & 55.18\% & 22.29\% \\
		DreamFly(Ours)	 	& 44.97m & 31.46\% & 46.77\% & 27.17\% \\
		\bottomrule
	\end{tabular}
\end{table}

\subsection{Qualitative Analysis}

\begin{figure}
	\centering
	\includegraphics[width=1.0\linewidth]{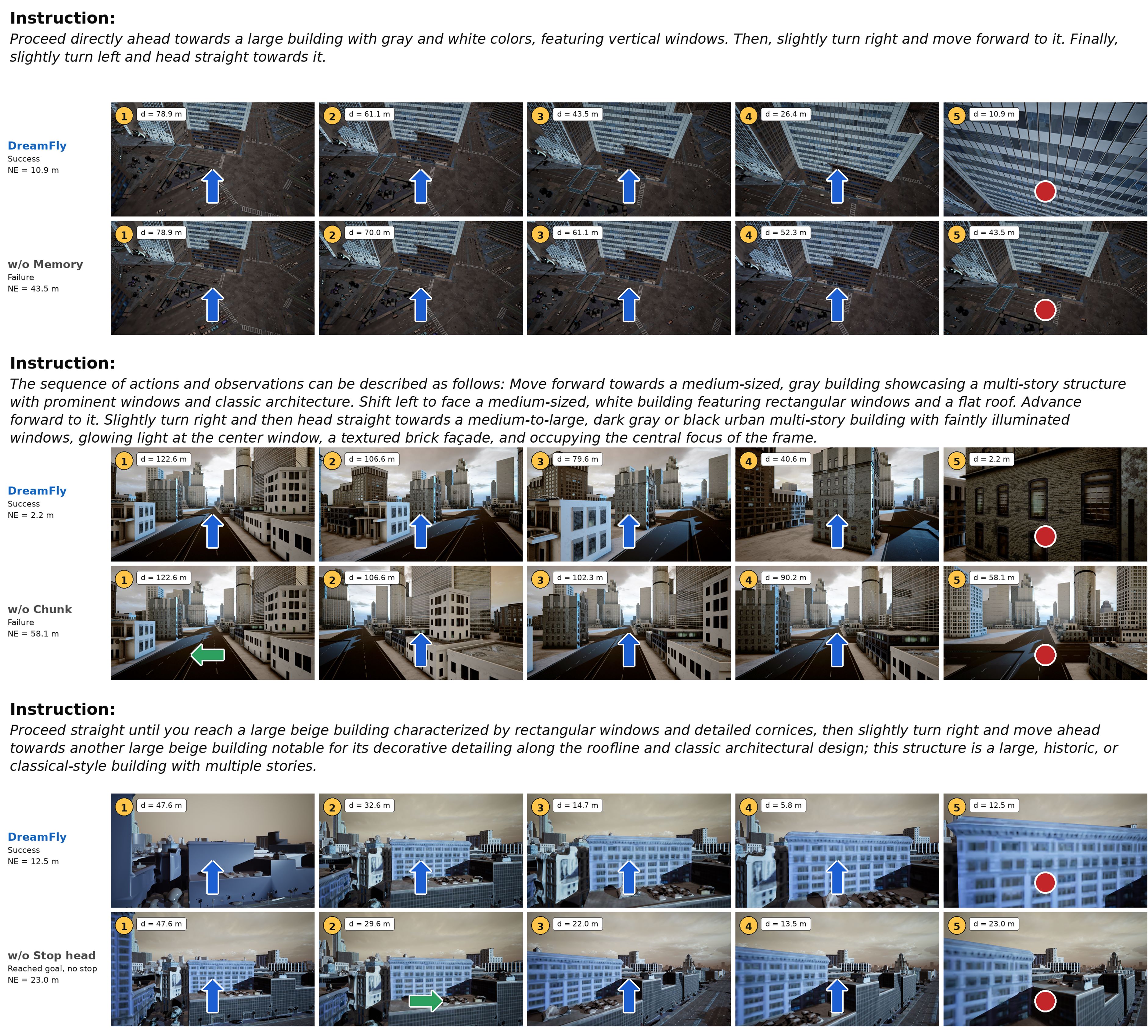}
	\caption{
		Qualitative comparison between DreamFly and its ablated variants on representative OpenFly trajectories. 
		The three examples illustrate the effects of historical memory, receding-horizon action planning, and LiteStop, respectively. 
		Numbers indicate the distance to the navigation goal.
	}
	
	\label{fig:qualitative}
\end{figure}

Fig.~\ref{fig:qualitative} qualitatively illustrates how the three proposed components affect closed-loop navigation. In the first example, DreamFly steadily approaches the target and reduces the navigation error from 78.9\,m to 10.9\,m, whereas removing historical memory results in much slower progress and a final error of 43.5\,m. This comparison shows that retaining previously observed visual evidence helps maintain a consistent reference to the target under changing viewpoints. In the second example, DreamFly follows a coherent sequence of actions and reaches the target with an error of 2.2\,m, while removing action-chunk prediction leads to an early deviation and eventually fails with an error of 58.1\,m, highlighting the benefit of incorporating short-horizon future-action structure into the current decision. In the third example, both variants approach the target region, but the model without LiteStop continues moving after entering the success region and drifts away, ending at 23.0\,m. DreamFly instead terminates successfully at 12.5\,m, demonstrating the importance of explicitly modeling task completion.

Overall, the examples reveal complementary failure modes addressed by the proposed components: historical memory preserves cross-step visual context, receding-horizon planning improves action consistency, and LiteStop prevents unnecessary motion after reaching the goal region.

\section{Conclusion}

In this work, we present DreamFly, a diffusion-based framework for aerial vision-language navigation. Building upon the Dream-VLA backbone, we revisit aerial navigation from the perspective of temporal decision making and identify three essential capabilities: retaining informative historical observations, anticipating future actions while preserving closed-loop feedback, and reliably determining when navigation should terminate. To this end, DreamFly integrates causally aligned historical memory, receding-horizon diffusion planning, and LiteStop for explicit termination into a unified closed-loop framework. The historical memory provides the policy with accumulated visual evidence under a strict causal prefix constraint, while diffusion-based action-chunk prediction enables joint modeling of the current action and short-horizon future actions. By following a plan-$K$, execute-one strategy, DreamFly uses future-action predictions as planning variables while replanning after every executed action, thereby preserving closed-loop feedback. LiteStop further decouples termination from action generation by estimating the stop probability from the initial all-mask action logits.

Extensive experiments on OpenFly validate the effectiveness of the proposed framework. DreamFly achieves the best NE, SR, and SPL on both the test-seen and test-unseen splits, reaching 32.04\%/29.46\% SR and 28.22\%/23.54\% SPL, respectively. The consistent performance improvements across both splits further demonstrate that DreamFly remains effective when navigating previously unseen environments.

Despite the consistent improvements observed in simulation, the current evaluation of DreamFly is still limited to simulated environments. Future work will focus on deploying the framework on physical UAV platforms to assess its real-world navigation capability and robustness under sensing noise, environmental disturbances, and sim-to-real domain shifts.

\printbibliography

\end{document}